%% file: manu_arxiv.tex
\documentclass{article} % For LaTeX2e
\usepackage{iclr2025_conference,times}

\input{math_commands.tex}

\usepackage{hyperref}
\usepackage{url}

\title{Can a Single Tree Outperform an Entire Forest?}

\author{Qiangqiang Mao, Yankai Cao \thanks{ Corresponding author: yankai.cao@ubc.ca.} \\
University of British Columbia, Vancouver, BC, Canada
}

\usepackage{multirow}
\usepackage{longtable}
\usepackage[flushleft]{threeparttable}
\usepackage{multicol}
\usepackage{graphicx}
\usepackage{booktabs} % For prettier tables

\usepackage{bbm}

\usepackage{enumitem}

\usepackage[capitalize,noabbrev]{cleveref}
\usepackage{algorithm}
\usepackage{algorithmic}

\iclrfinalcopy % Uncomment for camera-ready version, but NOT for submission.
\begin{document}

\maketitle

\begin{abstract}
The prevailing mindset is that a single decision tree underperforms classic random forests in testing accuracy, despite its advantages in interpretability and lightweight structure. This study challenges such a mindset by significantly improving the testing accuracy of an oblique regression tree through our gradient-based entire tree optimization framework, making its performance comparable to the classic random forest. Our approach reformulates tree training as a differentiable unconstrained optimization task, employing a scaled sigmoid approximation strategy. To ameliorate numerical instability, we propose an algorithmic scheme that solves a sequence of increasingly accurate approximations. Additionally, a subtree polish strategy is implemented to reduce approximation errors accumulated across the tree. Extensive experiments on 16 datasets demonstrate that our optimized tree outperforms the classic random forest by an average of $2.03\%$ improvements in testing accuracy. %Moreover, hypothesis testing confirms that our single tree performs comparably to random forests with a statistically significant difference. 

%This iterative approximation strategy effectively balances approximation degree and differentiability compared to a standard sigmoid function simply used for soft approximation.

\end{abstract}

\section{Introduction}
\label{sec:intro}

The single decision tree attracts significant attention in machine learning primarily due to its inherent interpretability. Its transparent ``IF-THEN'' decision rules make it highly useful for tasks that require clear decision-making logic behind predictions. However, its adoption is often limited by lower testing accuracy, particularly when compared to tree ensemble methods like the classic random forests \citep{Breiman_2001_Random}. Random forests, built from multiple decision trees, are widely recognized for their superior testing performance over single tree models \citep{Tan_2006_Decision}, and are considered among the best models for accuracy \citep{Fernandez-Delgado_2014_we,Grinsztajn_2022_Why}. However, a classic random forest, which typically consists of hundreds of decision trees, diminishes—or even eliminates—the interpretability that a single decision tree provides. This trade-off between interpretability and accuracy has become widely accepted, fostering a common mindset that the classic random forest outperforms single decision trees in testing accuracy, though at the expense of interpretability.

Such a mindset forces resorting to random forests when high test accuracy is essential, even in cases where a lightweight structure and interpretability are also demanded. For instance, in embedded systems with limited hardware resources and power budgets, a lightweight algorithm like decision tree is ideally preferable due to its fewer parameters and less energy consumption \citep{Narayanan_2007_FPGA, Alcolea_2021_FPGA}; yet, in practice, the subpar performance of a single tree often compels a shift toward tree ensembles, such as random forest \citep{Elsts_2021_Are, VanEssen_2012_Accelerating}. This shift introduces two major issues: first, it significantly aggregates computational costs and memory consumption due to more parameters from multiple trees. Second, it sacrifices the interpretability, which is crucial in certain decision-support scenarios, such as piece-wise control law in explicit model predictive control \citep{Bemporad_2002_Model} and threshold-based well control optimization in subsurface energy management \citep{Kuk_2022_Optimal}. These two concerns can be effectively addressed by a single decision tree if it could match the accuracy of random forests.

Aiming at a single tree with higher accuracy and fewer parameters, the oblique decision tree, a pivot extension of the classic orthogonal decision tree, holds great potential. Oblique decision trees use linear combination of features to create hyperplane splits. When the underlying data distribution follows hyperplane boundaries, oblique decision trees tend to simply tree structures, generating smaller trees with higher accuracy \citep{Costa_2023_Recent}. Nevertheless, inducing oblique decision trees presents substantial computational challenges, owing to the innumerable linear combinations of features at each node \citep{Zhu_2020_Scalable}. Earlier works mainly focus on finding the optimal feature combinations at an individual node using greedy top-down algorithms, such as \texttt{CART-LC} \citep{Breiman_1984_Classification} and \texttt{OC1} \citep{Murthy_1994_System}. Besides, alternative methods rely on greedy orthogonal decision trees \texttt{CART} to induce oblique trees by rotating the feature space, exemplified by \texttt{HHCART} \citep{Wickramarachchi_2015_HHCART} and \texttt{RandCART} \citep{Blaser_2016_Random}. Despite their advancements, such greedy methods that focus on optimal splits at current nodes, might lead to suboptimal solutions due to the weaker splits at subsequent child nodes. Considering the optimization of splits at all nodes, \citet{Bertsimas_2017_Optimal} presented optimal decision tree method to formulate tree training as a mixed-integer programming (MIP) problem. However, the practical application of MIP-based methods often face challenges in scalability and computational efficiency, especially in optimal oblique decision trees where the search space is expanded. Recent efforts in optimal oblique trees \citep{Boutilier_2023_Optimal,Zhu_2020_Scalable} have been confined to classification tasks with a limited number of categorical prediction values. In contrast, addressing regression tasks with an infinite number of possible prediction values remains an extremely challenging task. In response to this limitation, the originators of MIP-based work further proposed an alternative local search method \texttt{ORT-LS} \citep{Dunn_2018_Optimal} for tasks that are unsolvable by MIP. However, \texttt{ORT-LS} still suffers from high computational costs and suboptimal accuracy, as observed in our comparative studies. 

In this work, we reformulate the training of an entire tree as an unconstrained optimization task, offering significant solvability advantages over MIP reformulations. This reformulation makes it easily solvable through exiting powerful frameworks of gradient-based optimization. Given the non-differentiability of indicator functions in hard splits, two intuitive solutions has been used in recent literatures: treating the gradient of those indicators as one via straight-through estimators \citep{Karthikeyan_2022_Learning, Marton_2023_Learninga} and approximating indicators with sigmoid functions \citep{Wan_2021_NBDT,Yang_2018_Deep,Frosst_2017_Distilling}. However, straight-through estimators may neglect crucial gradient information, resulting in suboptimal outcomes, as observed in both their work \citep{Marton_2023_Learninga} and our experiments. In light of those works with sigmoid approximation, two major concerns arise. Firstly, previous efforts predominantly focused on constructing ``soft'' decision trees \citep{Irsoy_2012_Soft}, characterized by soft splits and probabilistic predictions. Nonetheless, there do exist scenarios where a hard-split tree with deterministic predictions is not only appropriate but also imperative. Further, the probabilistic soft splits significantly deviates from the interpretable True-False, IF-THEN decision logic. Secondly, the simple use of sigmoid functions leaves a considerable gap from indicator functions, necessitating a delicate balance between approximation accuracy and numerical solvability by scaling the sigmoid function \citep{Hehn_2017_End-to-end}. However, identifying the optimal scale factor also remains a challenge. More importantly, less attention has been paid to the approximation error that can accumulate across the entire tree, particularly in deep trees with numerous nodes. These issues substantially degrade the testing accuracy of a gradient-based tree, far lagging behind the performance of random forests. %These concerns highlight the need for further exploration into enhancing the accuracy of a single tree through gradient-based optimization.
% to make them competitive with random forests.

\textbf{Our contributions:} Firstly, we propose a strategy of iterative scaled sigmoid approximation to narrow the gap between the original indicator function and its differentiable approximation. This strategy uses the solution from an optimization task with a smaller scale factor to effectively warm-starts optimization with a larger scale factor. By starting with a smaller, smoother scale factor and gradually increasing it, this strategy enhance the approximation degree, while mitigating numerical instability typically associated with larger scale factors. Secondly, unlike soft trees with probabilistic predictions, we remain the hard-split decisions and deterministic predictions, only using soft approximation for gradient computations. Thirdly, to address severe approximation errors accumulated across each split in the entire tree, we propose a subtree polish strategy to further improve the training optimality. Finally, we provide an extensible \textbf{G}radient-based \textbf{E}ntire \textbf{T}ree optimization framework for inducing a tree with both constant predictions (termed as \textbf{\texttt{GET}}) and linear predictions (termed as \texttt{GET-Linear}), easily implemented in existing deep learning frameworks, as available in \url{https://github.com/maoqiangqiang/GET}.

\textbf{Performance:} Experiments show that our method can produce a tree with testing accuracy comparable to, or even exceeding, that of the classic random forest, challenging the prevailing mindset.
\begin{itemize}[itemsep=0pt,topsep=0pt,leftmargin=*]
   \item Empirically and statistically, our oblique tree \texttt{GET} outperforms compared decision tree methods  in test accuracy. Notably, it outperforms \texttt{CART} by 7.59\% and the state-of-the-art \texttt{ORT-LS} by 3.76\%.
   \item Our method \texttt{GET} statistically confirms its testing accuracy as comparable to classic random forest, and empirically underperforms the random forest by a mere 0.17\% gap.
   \item Our optimized oblique tree with linear predictions \texttt{GET-Linear} impressively outperforms random forests by an average of 2.03\% in testing, demonstrating a statistically significant difference.

\end{itemize}

\section{Foundations of Oblique Regression Tree}
\label{sec:obliquetree}

In this section, we explore oblique regression trees from an optimization perspective by formulating tree training as an optimization problem. 
For ease of understanding, we primarily follow the notation for optimal decision trees as used in the original work of \citet{Bertsimas_2017_Optimal}.

Consider a dataset comprising $n$ samples denoted as $\{\vx_i, y_i\}_{i=1}^n$ with input vectors $\vx_i \in [0, 1]^p$ and true output values $y_i \in [0, 1]$. 
% A min-max normalization procedure is applied to each sample for normalized input vectors $\vx_i \in [0, 1]^p$ and corresponding true output values $y_i \in [0, 1]$. 
A binary tree of depth $D$ comprises $T=2^{D+1}-1$ nodes, where each node is indexed by $t \in \sT=\{1,\cdots,T\}$ in a breadth-first order. 
The nodes can be categorized into two types: branch nodes, which execute branching tests and are denoted by indices $t\in \sT_B = \{1,\cdots,\lfloor T/2 \rfloor\}$, and leaf nodes denoted by $t\in \sT_L = \{\lfloor T/2 \rfloor + 1,\cdots, T\}$, responsible for providing regression predictions. 
% The nodes can be categorized into two distinct classes: branch nodes, which execute branching tests and are denoted by indices $t\in \sT_B = \{1,\cdots,\lfloor T/2 \rfloor\}$, and leaf nodes, responsible for providing regression predictions and identified by indices $t\in \sT_L = \{\lfloor T/2 \rfloor + 1,\cdots, T\}$. 
Each branch node comprises a split weight $\va_t\in \sR^p$ 
% Each branch node comprises a split weight $\va_t = \{a_{t, 1}, \cdots, a_{t, p}\}\in \sR^p$ 
and a split threshold $b_t\in \sR$ to conduct a branching test ($\va_t^T \vx_i \leq b_t$) for the samples allocated to that particular branch node. If a sample $\vx_i$ passes the branching test ($\va_t^T \vx_i \leq b_t$), it is directed to the left child node at index $2t$; otherwise, to the right child node at index $2t+1$.
% If a sample $\vx_i$ successfully satisfies the branching test ($\va_t^T \vx_i \leq b_t$), it is directed to the left child node with index $2t$; conversely, if it fails the test, it is directed to the right child node with index $2t+1$. 
Each leaf node comprise the parameters of $\vk_t \in \sR^p$ and $h_t\in \sR$ to provide a prediction value for current leaf. The training of oblique regression trees involves solving the following optimization problem:
\begin{subequations}\label{eq:ortlossall}
	\begin{align}
      \min_{\mA, \vb, \mK, \vh} & \  \sum_{i=1}^n \left(y_i - \hat{y_i} \right)^2, \label{eq:ortlos} \\ 
       \ & \text{s.t.} \ \; \hat{y_i} = f_{tree}(\mA, \vb, \mK, \vh,\vx_i), \ i \in \{1, \cdots, n\} ,\label{eq:yihat}
    \end{align}
\end{subequations} 
where 
% the final prediction $\hat{y}_i = \vk_t^T\vx_i+h_t$ if $\vx_i$ is assigned to leaf note $t \in \sT_L$. 
$\mA = \{\va_1, \cdots, \va_{\lfloor T/2 \rfloor}\}$ and $\vb = \{b_1, \cdots, b_{\lfloor T/2 \rfloor}\}$ are tree split parameters for branch nodes, $\mK= \{\vk_{\lfloor T/2 \rfloor+1}, \cdots, \vk_T\}$ and $\vh= \{h_{\lfloor T/2 \rfloor+1}, \cdots, h_T\}$ are leaf prediction parameters. 
% The prediction parameters may slightly differ depending on the type of leaf prediction chosen. 
In this work, we consider two types of leaf prediction: \textbf{(a)} linear prediction and \textbf{(b)} constant prediction.  

\textbf{(a) Tree with linear predictions:} Linear predictions involve a linear combination of input features \citep{Quinlan_1998_Learning}, representing a general form of leaf predictions.  If $\vx_i$ is assigned to leaf node $t$, the final prediction is described as $\hat{y}_i = \vk_t^T\vx_i + h_t$.

\textbf{(b) Tree with constant predictions:} This type is a special case of linear predictions, where $\mK$ remains zero. It is the most commonly used type in existing decision tree methods, with $\hat{y}_i = h_t$.

\section{Unconstrained Optimization Formulation}
In this work, we reformulate the tree training as an unconstrained optimization task, allowing us to leverage powerful gradient-based optimization frameworks for improved solvability and accuracy.
\subsection{Deterministic Sample Route Formulation}
\label{sec:samplerouteformulation}
The interpretability of decision trees primarily stems from their transparent prediction rules associated with the tree paths from the root to leaf nodes. Identifying a sample's specific tree path, such as ``$1\rightarrow2\rightarrow5$'' for a sample routed to leaf node 5, is crucial for calculating prediction loss.
Specifically, for a leaf node $t \in \sT_L$, we denote its set of ancestor nodes as $\sA_t$. The subsets $\sA_t^l$ and $\sA_t^r$ represent the ancestor nodes traversed via the left branch and right branch, respectively, such that $\sA_t$ = $\sA_t^l \cup \sA_t^r$.
Additionally, we introduce a binary branching test variable $I_{i,j}\in \{0, 1\}$ to signify whether a sample $\vx_i$ successfully passes the branching test at the branch node $j$, defined as $I_{i, j}= \mathbbm{1}\left( b_{j}-\va_{j}^T \vx_i >0 \right)$. Here, $I_{i,j}=1$ signifies a successful pass at node $j$; otherwise, $I_{i,j}=0$.

\begin{figure}[!htp]
   % \vskip 0.2in
   \begin{center}
   \centerline{\includegraphics[width=0.5\textwidth]{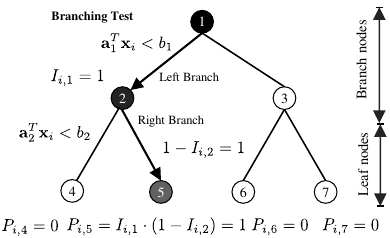}}
   % \vskip -0.1in
   \caption{Example of deterministic sample routing $1\rightarrow 2 \rightarrow 5$.}
   \label{fig:125}
   \end{center}
   %  \vskip -0.3in
\end{figure}

Subsequently, the sample routing indicator $P_{i,t}\in \{0,1\}$, determines if sample $\vx_i$ is assigned to leaf node $t$, computed as follows:
\begin{equation} \label{eq:pt}
   P_{i,t} = \prod_{j\in \sA_t^l} I_{i,j} \prod_{j\in \sA_t^r} \left(1-I_{i,j}\right),
\end{equation}
where $P_{i,t}=1$ indicates an assignment to the leaf node $t$, and $P_{i,t}=0$ denotes non-assignment. An illustrative example for understanding these formulations is provided in \cref{fig:125}.
% that sample $\vx_i$ is assigned to the leaf node $t$, while $P_{i,t}=0$ signifies that sample $\vx_i$ is not assigned to the leaf node $t$. An illustrative example for understanding these formulations is provided in \cref{appendix:125example}, \cref{fig:125}.
We use a 2-depth tree, where the leaf nodes are indexed by $\sT_L={4, 5, 6, 7}$. Focusing on leaf node $t=5$, the corresponding tree path is delineated as "$1 \rightarrow 2 \rightarrow 5$". The associated ancestor sets are specified as $\sA_5^l={1}$ and $\sA_5^r={2}$. Assuming the $i$th sample is assigned to leaf node 5 by successfully passing the branching test on branch node 1 with $\va_1^T \vx_i < b_1$ and failing on branch node 2 with $\va_2^T \vx_i \geq b_2$, the ensuing outcomes are $I_{i,1}=1$, $I_{i,2}=0$, $P_{i,5} = I_{i,1} \cdot (1-I_{i,2}) = 1$, and $P_{i,4}=P_{i,6}=P_{i,7}=0$. Consequently, this signifies that the sample $\vx_i$ is assigned to leaf node 5. 

\subsection{Loss Formulation and Differentiability}
\label{sec:lossformulation}
Following the deterministic sample route, the training of oblique regression trees is subsequently reformulated as an unconstrained optimization problem. The objective function $\mathcal{L}$ is defined by 
\begin{equation}
   % \mathcal{L} = \sum_{i=1}^{n}\sum_{t\in\mathcal{T}_L} \left[\prod_{j\in \mathcal{A}_t^l} I_{i,j} \prod_{j\in \mathcal{A}_t^r} \left(1-I_{i,j}\right)\right] (y_i-c_t)^2. \label{eq:loss}
   \mathcal{L} = \sum_{i=1}^{n}\sum_{t\in\sT_L} P_{i,t} \left(y_i-(\vk_t^T\vx_i+h_t)\right)^2. \label{eq:loss}
   % \mathcal{L} = \sum_{i=1}^{n}\sum_{t\in\sT_L} \left[\prod_{j\in \mathcal{A}_t^l} I_{i,j} \prod_{j\in \mathcal{A}_t^r} \left(1-I_{i,j}\right)\right] (y_i-(\vk_i^T\vx_i+h_t))^2. \label{eq:loss}
\end{equation}
Here, the variables $\mA$, $\vb$, $\mK$ and $\vh$ are implicitly expressed in terms of $I_{i,j}$ and $P_{i,t}$, as shown in \cref{eq:pt}. For decision trees with constant predictions, the variable $\mK$ is always equal to zero. 

\begin{figure}[!htp]
    % \vskip 0.2in
    \begin{center}
    \centerline{\includegraphics[width=0.8\columnwidth]{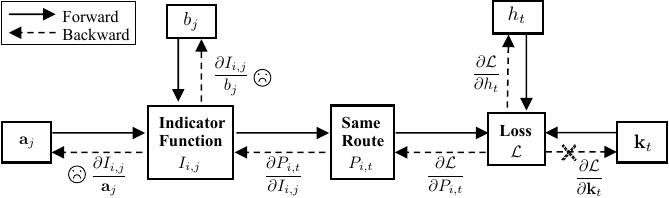}}
    % \vskip -0.1in
    \caption{Forward and backward processes of loss calculation.}
    \label{fig:grad}
    \end{center}
   %  \vskip -0.3in
 \end{figure}

In the computation of gradients of $\mathcal{L}$ (as detailed in \cref{fig:grad}), an exception arises due to the non-differentiability of the indicator function $\mathbbm{1}(\cdot)$ in the calculation of the branching test variable $I_{i,j}$. To resolve this issue, we employ the scaled sigmoid function $\mathbb{S}(\cdot)$ as an approximation for the indicator function, resulting in the introduction of the approximated branching test variable denoted as $\hat{I}_{i,j}$:
\begin{equation}\label{eq:hatsigmoid}
   % S_{i,j}  = \frac{1}{1+e^{-\alpha \left(b_j-\mathbf{a}_j^T \mathbf{x}_i\right)}}
   % S_{i,j}  = \left[1+e^{-\alpha \left(b_j-\mathbf{a}_j^T \mathbf{x}_i\right)}\right]^{-1}
   \hat{I}_{i, j}= \mathbb{S}(b_j-\mathbf{a}_j^T \mathbf{x}_i) = \left[1+e^{-\alpha \left(b_j-\mathbf{a}_j^T \mathbf{x}_i\right)}\right]^{-1},
\end{equation}
where $\alpha$ represents a critical balance between achieving high approximation accuracy and maintaining stability in optimization processes, while also providing some potential concerns.

\textbf{Concerns regarding the special case:} $\alpha=1$ corresponds to standard sigmoid function, commonly used in Soft Decision Tree \citep{Irsoy_2012_Soft,Frosst_2017_Distilling}, which adopt soft splits at branch nodes and probabilistic predictions at leaf nodes. Their work on soft trees deviates from the interpretability pertinent to hard True-False decision and deterministic predictions, making it less suitable for scenarios requiring hard-splits, as exemplified in \cref{appendix:scenarios}. More importantly, there remains a big gap between the standard sigmoid function with $\alpha=1$ and the true indicator function, diminishing the accuracy of the training.

\textbf{Concerns regarding the selection of $\alpha$:} A larger $\alpha$ leads to a more accurate approximation, but it also introduces numerical instability, potentially compromising the optimization capabilities. Further empirical analyses are given in \cref{appendix:alpha}. Identifying the optimal $\alpha$ that balances approximation degree and differentiability remains a challenge. To mitigate this, we propose an iterative scaled sigmoid approximation strategy, detailed in the following \cref{subsec:iterative}, to narrow the gap between the original indicator function and its differentiable approximation.

\textbf{Clarification for the adoption of hard-split in inference:} Unlike soft trees that base final predictions on probability and trained leaf values ($\mK$ and $\vh$), our tree still maintains hard-splits. Hard splits in the inference phase not only meet the practical demand for hard decisions but also mitigates additional errors that may arise from the trained leaf values. Ideally, with a high approximation accuracy and optimal optimization, the trained leaf values from soft approximations should closely match those calculated by hard splits. However, achieving this level of optimality is challenging due to potential errors in soft approximation. This concern justifies the use of hard splits in the inference phase, which are more likely to yield more accurate final predictions.

\section{Oblique Tree Training Through Gradient-based Optimization}
Our optimization task in \cref{eq:loss}, incorporating the approximated $\hat{I}_{i,j}$ obtained from \cref{eq:hatsigmoid}, closely approximate the original non-differentiable tree training problem. This task can be efficiently solved through our proposed entire tree optimization framework. 

\subsection{Iterative Scaled Sigmoid Approximation}
\label{subsec:iterative}
In response to early-mentioned concerns regarding an optimal $\alpha$, we propose a strategy of iterative scaled sigmoid approximation to enhance the approximation accuracy to indicator functions. The key challenge of lies in the selection of $\alpha$. A larger $\alpha$ may destabilize optimization process, whereas a smaller $\alpha$ tends to be easier to solve for gradient-based optimization. Our strategy leverage this insight by using a solution from an optimization task with a smaller scale factor to effectively warm-starts optimization with a larger scale factor. By starting with a smaller scale factor and gradually increasing it, this strategy enhance the approximation accuracy, while mitigating numerical instability typically associated with larger scale factors. 

Specifically, the procedure begins by randomly sampling a set of scale factors within a predetermined range, ensuring a broad exploration of possible $\alpha$ values. These sampled scale factors, denoted as $\{\alpha_1, \cdots, \alpha_n\}$, are then applied in ascending order, from small to large. We initiate the optimization with the smallest sampled scale factor to generate the initial optimized tree candidate. This candidate then serves as the starting point for the subsequent optimization task with a slightly larger $\alpha$. This iterative process is repeated until all sampled scale factors have been utilized. Detailed implementation steps are integrated within our systematic optimization framework, \cref{alg:GET}.

\subsection{Gradient-based Entire Tree Optimization Framework}
\label{subsec:entiretree}
Unlike greedy methods that optimize each node sequentially, our approach concurrently optimizes the entire tree, encompassing tree split parameters $\mA$ and $\vb$ at all branch nondes, and leaf prediction parameters $\mK$ and $\vh$ at all leaf nodes. Our entire tree optimization, outlined in \cref{alg:GET}, begins at multiple random initialization (\textit{Line~\ref{line:start}~-~\ref{line:alphaSet}}). This multiple-initialization serves two purposes: First, multi-start increases the chance of finding the optimal solution of the unconstrained reformulation. Second, in \cref{subsec:iterative}, the iterative scaled sigmoid approximation involves randomly sampling scale factors for each iteration. These multiple starts enable sampling diverse scale factors in a wider range, thereby enhancing the robustness and approximation accuracy. For each start, the optimization with iterative scaled sigmoid approximation is implemented to produce an optimized tree candidate (\textit{Line~\ref{line:alphaIter}~-~\ref{line:return}}). Importantly, our method deterministically calculates the leaf values based on hard-splits (\textit{Line~\ref{line:kupdate}}). The specific deterministic calculations for $\mK$ and $\vh$ are given in \cref{appendix:deterministicCal}. Finally, the optimal tree is determined by comparing each candidate from multiple starts (\textit{Line~\ref{line:lossupdate}~-~\ref{line:compareloss}}). This optimization framework is readily implementable using existing powerful tools that embed gradient-based optimizers, such as PyTorch and TensorFlow. 

The framework is applicable to decision trees with both constant and linear predictions. The only minor difference is that for constant predictions, the parameters $\mK$ remain zero without gradients. For clarity, we term our approach \textbf{G}radient-based \textbf{E}ntire \textbf{T}ree optimization as \textbf{\texttt{GET}} when applied to trees with constant predictions; otherwise, termed as \texttt{GET-Linear} for trees with linear predictions.

\begin{algorithm}[!htp]
   \small
   \caption{Gradient-based Entire Tree Optimization (applicable to both \texttt{GET} and \texttt{GET-Linear})}
   \label{alg:GET}
   \begin{algorithmic}[1]
      \STATE {\bfseries Input:} $\{\vx_i, y_i\}_{i=1}^n$, tree depth $D$, learning rate $\eta$, epoch number $N_{epoch}$, multi-start number $N_{start}$.
      \STATE {\bfseries Output:} Optimal trainable variables $\mA_{best}$, $\vb_{best}$, $\mK_{best}$ (Zero for the case of \texttt{GET}) and $\vh_{best}$.
      % \STATE Initialize trainable variables $\mathbf{A}$, $\mathbf{b}$ and $\mathbf{c}$.
      \STATE Assign a large value to the parameter $\mathcal{L}_{min}$ and define empty variables for $\mA_{best}$, $\vb_{best}$, $\mK_{best}$ and $\vh_{best}$. 
      \FOR{$start=1 $ {\bfseries to} $N_{start}$} \label{line:start}
         \STATE Initialize trainable variables $\mA$, $\vb$, $\mK$ (Zero for \texttt{GET}) and $\vh$. \label{line:init}
         \STATE Randomly generate a set of scale factors $\{\alpha_1, \cdots, \alpha_n\}$ in ascending order. \label{line:alphaSet}
         \FOR{$\alpha_{iter} \in \{\alpha_1, \cdots, \alpha_n\}$} \label{line:alphaIter}
            \STATE If $iter \neq 1$, initialize trainable variables from the solution of last iteration computed by \textit{Line~\ref{line:Aiter}}.
            % \STATE Initialization by last tree candidate from a smaller scale by \textit{Line~\ref{line:Aiter}} IF $\alpha \neq \alpha_1$ ELSE by \textit{Line~\ref{line:init}}.
            \FOR{$k=1$ {\bfseries to} $N_{epoch}$}
               \STATE Approximate loss $\mathcal{L}$ at step $k$ by \cref{eq:loss} and (\ref{eq:hatsigmoid}) and calculate $\frac{\partial \mathcal{L}}{\partial \mA}$, $\frac{\partial\mathcal{L}}{\partial \vb}$, $\frac{\partial \mathcal{L}}{\partial \mK}$ (exclude for \texttt{GET}) and $\frac{\partial \mathcal{L}}{\partial \vh}$. Then update trainable variables, such as $\mA_{k+1} = \mA_{k} - \eta \frac{\partial \mathcal{L}}{\partial \mA}$.
            \ENDFOR
            \STATE Deterministically update $\mK$ and $\vh$ based on hard-splits.  \label{line:kupdate}
            \STATE Generate a tree candidate with optimized variables, termed as $\mA_{iter}$, $\vb_{iter}$, $\mK_{iter}$ and $\vh_{iter}$. \label{line:Aiter}
            \STATE Deterministically compute the current loss $\mathcal{L}$ based on hard-splits using \cref{eq:loss}.\label{line:lossupdate}
            \STATE IF {$\mathcal{L} < \mathcal{L}_{min}$}, $\mA_{best} \gets \mA_{iter}$; $\vb_{best} \gets \vb_{iter}$; $\mK_{best} \gets \mK_{iter}$; $\vh_{best} \gets \vh_{iter}$; $\mathcal{L}_{min} \gets \mathcal{L}$. \label{line:compareloss}
         \ENDFOR \label{line:alphaIterEnd}
      \ENDFOR \label{line:return}
      % \STATE \textbf{Return} optimal trainable variables $\mA_{best}$, $\vb_{best}$, $\mK_{best}$, and $\vh_{best}$. \label{line:return}
   \end{algorithmic}
\end{algorithm}
% \vskip -0.5in

\textbf{Hyperparameters analysis:} Despite the introduction of additional hyperparameters in gradient-based optimization, tuning them is not typically necessary because their effects are straightforward. For instance, the multi-start number $N_{start}$ directly influences training optimality by increasing the chance of finding optimal solution, albeit at a higher computational cost. In practice, $N_{start}$ is set to balance acceptable computational cost with desired training accuracy, and a similar approach is applied to $N_{epoch}$. More empirical analyses of $N_{start}$, $N_{epoch}$, the sampling range of scale factors and other hyperparameters are detailed in \cref{appendix:hyperparameters}.

\subsection{Subtree Polish Strategy to Mitigate Accumulated Approximation Errors} 
\label{subsec:subtree}
\textbf{Accumulated approximation error analysis:} Despite our approximation strategy effectively narrowing the substantial gap to the indicator function at each node, the approximation error can still accumulate across an entire decision tree, particularly in deeper trees with numerous nodes. This accumulation can significantly degrade the training optimality, a concern that has received insufficient attention in the literature. To mitigate these accumulated errors, we propose a subtree polish strategy to further enhance training optimality.

\begin{figure}[!htp]
   % \vskip -0.1in
   \begin{center}
   \centerline{\includegraphics[width=\textwidth]{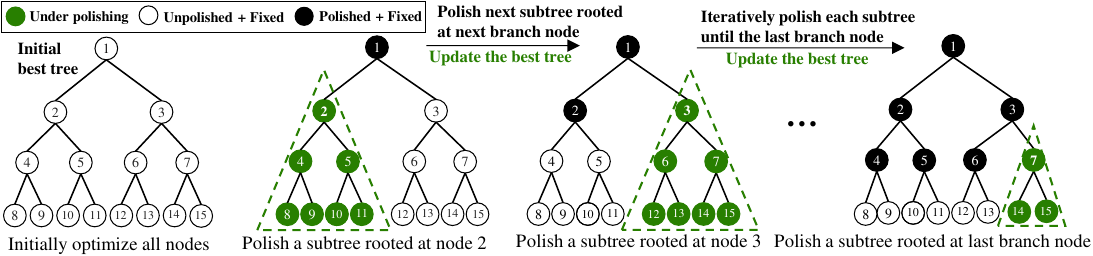}}
   \vskip -0.1in
   \caption{The illustrative example of our subtree polish strategy.}
   \label{fig:subtree}
   \end{center}
   % \vskip -0.3in
\end{figure}

\textbf{Subtree polish strategy to improve training optimality:} In \cref{alg:GET}, Our method is designed to simultaneously optimize all $2^D-1$ branch nodes of an entire tree. Intuitively, once a branch node is optimally identified, we can sequentially polish its subtree, including all child nodes of that branch node, while keeping the rest of the tree fixed. This process forms the basis of our subtree polish strategy, which starts with optimizing the entire tree, establishing an initial best tree candidate. 

For each branch node, there exists a corresponding subtree that extends from the current branch node to the leaf nodes. As illustrated in \cref{fig:subtree}, the subtree is represented within a dashed triangle. Each subtree is optimized using the same approach as for entire tree in \cref{alg:GET}, while leaving the remaining tree nodes fixed. Each subtree optimization is warm-started with the best tree candidate available at that time. Once a subtree is optimized, we combine it with the fixed tree nodes and update the best tree if the combined tree improves training accuracy. This process then proceeds to next subtree, rooted at next branch node. This iterative process continues until the subtree rooted at the last branch node is polished. The final best tree is returned after polishing all branch nodes. The implementation procedure of the subtree polish strategy is provided in \cref{appendix:subtreepolish}, \cref{alg:subtreepolish}.

\section{Numerical Experiments and Discussions}
\label{sec:experiments}
We evaluate our optimized tree with both constant and linear predictions, termed as \texttt{GET} and \texttt{GET-Linear}, against the classic random forests \texttt{RF} \citep{Breiman_2001_Random}. Our analysis covers testing accuracy, the number of parameters and prediction time. Additionally, we assess the capabilities of \texttt{GET}, with other decision tree methods in terms of both training optimality and testing accuracy. Finally, we provide a limitation analysis. The compared tree methods include the baseline \texttt{CART}, greedy methods like \texttt{HHCART} \citep{Wickramarachchi_2015_HHCART}, \texttt{RandCART} \citep{Blaser_2016_Random}, and \texttt{OC1} \citep{Murthy_1994_System}, existing gradient-based trees such as \texttt{GradTree} \citep{Marton_2023_Learninga} using a straight-through estimator for non-differentiable splits, and soft decision tree \texttt{SoftDT} \citep{Frosst_2017_Distilling} using standard sigmoid function for soft approximation, as well as the state-of-the-art heuristic method \texttt{ORT-LS} \citep{Dunn_2018_Optimal}. These comprehensive experiments are conducted on 16 real-world datasets obtained from UCI machine learning repository \citep{Dua_2019_UCI} and OpenML \citep{Vanschoren_2014_OpenML}, with sample size ranging from 1,503 to 16,599 and feature number from 4 to 40. Detailed dataset information and specific data usage in our study is provided in \cref{appendix:dataset}. Comparisons focus on testing and training accuracy in terms of Coefficient of Determination (usually denoted as $R^2$), and computational time in seconds. The Friedman Rank \citep{Sheskin_2020_Handbook} is also used to statistically sort the compared methods according to their testing accuracy, with a lower rank indicating better performance. Comprehensive details on the implementation, algorithm configuration, and computing facilities are provided in \cref{appendix:implementation}.

\subsection{Testing Accuracy Comparison Against Random Forests}
\label{subsec:ComprVSRF}
% To showcase the superior predictive accuracy of our optimized tree, we compare the testing accuracy of our methods, \texttt{GET} for trees with constant predictions and \texttt{GET-Linear} for trees with linear predictions, against Random Forests (\texttt{RF}). 
For testing accuracy comparison, we conduct depth tuning by cross validation to determine the optimal depth across depths from 1 to 12 for our methods. For a fair comparison, comprehensive hyperparameter tuning is also performed for \texttt{RF}. Specifically, the number of trees in a forest is a critical parameter. It is well-recognized that testing performance improves with an increase in the number of trees; however, the marginal gains become less pronounced as additional trees are added \citep{Probst_2018_Tune, Oshiro_2012_How}. Accordingly, the number of trees for \texttt{RF} is tuned across a set of $\{50, 100, 200, 300, 400, 500\}$. Moreover, the maximum tree depth for \texttt{RF} is tuned over a broader range, from 1 to 50, to potentially capture optimal depth settings, given that \texttt{RF} empirically benefits from overly-deeper trees for enhanced testing accuracy. Other hyperparameters for \texttt{RF}, including the number of features per split and the number of samples per tree, are maintained at default settings. These parameters have been shown to balance the bias-variance trade-off, typically yielding robust performance with default values \citep{Probst_2018_Tune}. 

\textbf{Testing accuracy comparison:} Our \texttt{GET} slightly underperforms \texttt{RF} by 0.17\% averaging across 16 datasets. In contrast, \texttt{GET-Linear} significantly outperforms \texttt{RF} by 2.03\%, as detailed in \cref{tab:TestVSRF}. This superiority is further supported by the Friedman Rank, where \texttt{GET-Linear} ranking the highest, followed by \texttt{RF} and \texttt{GET}. Among 16 datasets, \texttt{GET-Linear} outperforms \texttt{RF} in 8 datasets. Detailed results for each dataset are given in \cref{appendix:testComparison}. These findings question the conventional belief that a single decision tree typically underperforms random forests, highlighting the capabilities of our approach in achieving competitive testing accuracy.

\begin{table}[!htp]
   \tabcolsep=12pt
   % \renewcommand{\arraystretch}{0.1}
   % \vskip -0.15in
   \centering
   \caption{Comparison of testing accuracy for \texttt{GET}, \texttt{GET-Linear}, and \texttt{RF} across 16 datasets.}
   \label{tab:TestVSRF}
   \resizebox{\columnwidth}{!}{%
   \begin{tabular}{ccccc}
      \toprule
      Item       & \multicolumn{1}{l}{Number of Trees} & Tree Depth & Test Accuracy ($R^2$, \%) & Friedman Rank \\ \hline
      \texttt{RF}         & 309.38                              & 19.24 & 81.94                 & 1.75 \\
      \texttt{GET}        & 1                                & 6.56  & 81.77                 & 2.69 \\
      \texttt{GET-Linear} & 1                                & 6.94  & \textbf{83.97}        & 1.56 \\ \bottomrule
      \end{tabular}%
   }
   % \vskip -0.05in
   \end{table}

\textbf{Paired T-test for statistical significance:} While \cref{tab:TestVSRF} empirically shows that our single tree is competitive to \texttt{RF}, we further conduct a Paired T-test to statistically validate the significant difference among \texttt{GET}, \texttt{GET-Linear} and \texttt{RF}. The null hypothesis asserts that \texttt{RF} is not significantly different from \texttt{GET} and \texttt{GET-Linear}. By setting a tolerance (acceptable significance level) $\tau=0.1$, if calculated $p$-value is less than $\tau$, the null hypothesis can be rejected, indicating statistical significance.

\begin{figure}[!htp]
   % \vskip -0.15in
   \begin{center}
   \centerline{\includegraphics[width=0.5\textwidth]{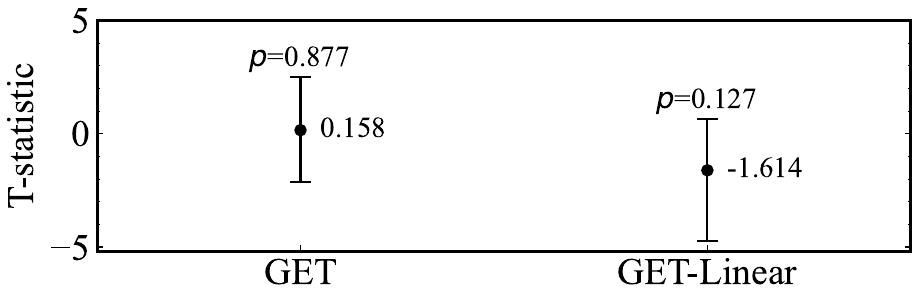}}
   % \vskip -0.1in
   \caption{T-statistic and p-value of the Paired T-test comparing \texttt{RF} with our methods.}
   \label{fig:ttestRF}
   \end{center}
   % \vskip -0.15in
\end{figure}

The t-statistic (black point), p-value, 95\% confidence interval are depicted in \cref{fig:ttestRF}. The Paired T-test between \texttt{RF} and our \texttt{GET} yields a $p$ value of 0.877 (greater than $\tau$), suggesting no statistically significant difference in testing accuracy between \texttt{RF} and \texttt{GET}. Thus, their testing accuracies are comparable. In contrast, the $p$ value for T-test between \texttt{RF} and \texttt{GET-Linear} is relatively smaller with $p=0.127$, also suggesting that \texttt{RF} is not significantly better than \texttt{GET-Linear}. Moreover, if we accept a tolerance $\tau>0.127$, we can reject the null hypothesis of equal performance. As indicated by a negative t-statistic of -1.164, this implies that \texttt{GET-Linear} is statistically superior to \texttt{RF}. These results further confirm the competitive testing accuracy of our tree over random forests.

\subsection{Testing Accuracy Comparison Against Other Decision Trees}
Following our empirical findings, which showcases the superiority of our optimized tree, we proceed to compare it with other commonly-used decision tree methods under same depth tuning setting, to further validate the efficacy of our approach. For a fair comparison, since existing decision trees are mainly designed with constant predictions, we limit the comparison to our \texttt{GET} with them. 

\begin{table}[!htp]
   \centering
   % \vskip -0.2in
   \caption{Comparison of test accuracy for \texttt{GET} and other decision tree methods.}
   \label{tab:testVSDT}
   \resizebox{\columnwidth}{!}{%
   \begin{tabular}{c|cccc|cc|c|c}
   \hline
   \multirow{2}{*}{Item} & \multicolumn{4}{c|}{Greedy Methods} & \multicolumn{2}{c|}{Gradient-based Trees} & State-of-the-Art Heuristic & Our Tree       \\ \cline{2-9} 
    &
     \texttt{CART} &
     \texttt{OC1} &
     \texttt{RandCART} &
     \texttt{HHCART} &
     \texttt{SoftDT} &
     \texttt{GradTree} &
     \texttt{ORT-LS} &
     \texttt{GET} \\ \hline
   Test Accuracy ($R^2$, \%)    & 74.18   & 72.54   & 71.31  & 76.37  & 72.69               & 64.13               & 78.01                      & \textbf{81.77} \\
   Tree Depth            & 10.06   & 7.94    & 8      & 8.19   & 10.19               & 10.38               & 6.13                       & 6.56           \\
   Friedman Rank         & 4.81    & 5.19    & 5.88   & 3.63   & 4.69                & 6.94                & 3.50                       & \textbf{1.38}  \\ \hline
   \end{tabular}%
   }
   \end{table}

\textbf{Testing accuracy comparison:} Our \text{GET} consistently outperform compared decision trees in testing accuracy across 16 datasets, as shown in \cref{tab:testVSDT}. Specifically, \texttt{GET} achieves the highest testing accuracy, surpassing the state-of-art heuristic method \texttt{ORT-LS} by 3.76\%, the greedy method \texttt{HHCART} by 5.39\%, and the baseline orthogonal tree \texttt{CART} by 7.59\%. Notably, compared to other gradient-based trees, \texttt{GET} also outperforms \texttt{GradTree} by 17.64\%, and the soft decision tree \texttt{SoftDT} by 9.08\%. These results underscore the effectiveness of our tree optimization approach. Additionally, the rank comparisons reinforce these findings, with \texttt{GET} attaining the highest rank. Detailed results for each dataset are provided in \cref{appendix:testComparison}.

\textbf{Paired T-test for statistical significance:} The Paired T-test for all comparisons consistently shows that we can reject the null hypothesis, which posits that our method \texttt{GET} is not significantly different from the compared decision trees. The observations of $p<\tau=0.1$ and positive t-statistic value, indicate that \texttt{GET} is superior to these compared decision trees with statistically significance. 

\begin{figure}[!htp]
   % \vskip -0.08in
   \begin{center}
   \centerline{\includegraphics[width=0.7\textwidth]{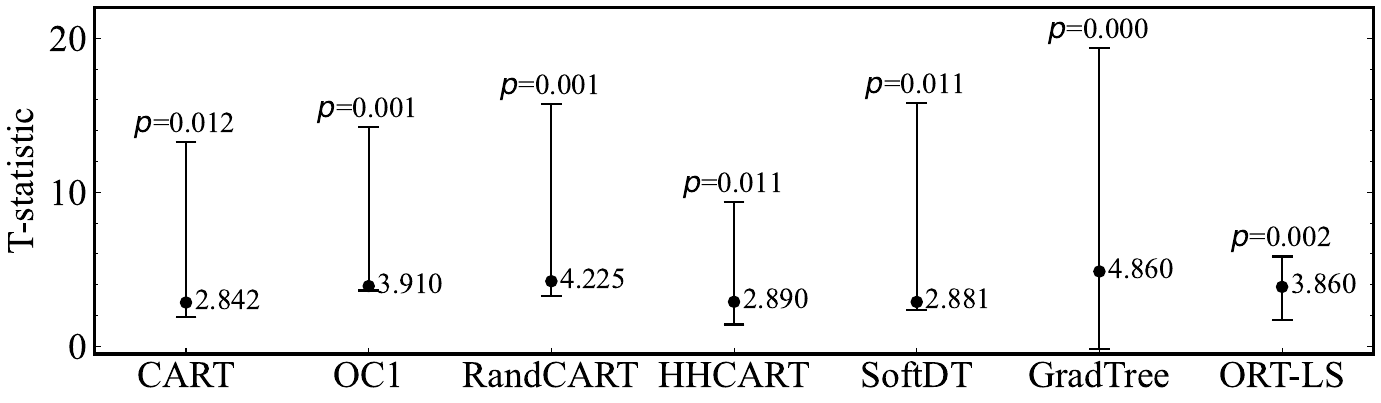}}
   \vskip -0.07in
   \caption{The Paired T-test comparing our \texttt{GET} with various decision tree methods.}
   \label{fig:ttestDT}
   \end{center}
   % \vskip -0.29in 
\end{figure}

\subsection{Superior Test Accuracy Analysis: From Training Optimality Perspective}
To figure out the rationale behind the competitive testing accuracy of our optimized tree, we delve into the analysis of training optimality from optimization perspective. Notably, the training of tree with different depth corresponds to different optimization problems, exhibiting different optimization challenge in scalability and performance. To assess the optimization capabilities, the training of a predetermined-depth tree is a common practice in the literature. Therefore, the predetermined depths of $D =\{2,4,8,12\}$ are used for training accuracy comparison. 
Detailed results for all compared decision tree methods in terms of training accuracy, testing accuracy and training time are given in \cref{tab:TrainforAllMethods}.

\begin{table}[!htp]
    \centering
    \caption{Detailed results for all compared decision tree methods under fixed depth setting.}
    \label{tab:TrainforAllMethods}
    \resizebox{\columnwidth}{!}{%
    \begin{tabular}{c|c|cccc|cc|c|c}
    \toprule
    \multirow{2}{*}{Item} &
      \multirow{2}{*}{D} &
      \multicolumn{4}{c|}{Greedy Methods} &
      \multicolumn{2}{c|}{Gradient-based Trees} &
      State-of-Art Heuristic &
      Our Tree \\ \cline{3-10} 
     &    & \texttt{CART}  & \texttt{OC1}     & \texttt{RandCART} & \texttt{HHCART} & \texttt{SoftDT}    & \texttt{GradTree} & \texttt{ORT-LS}         & \texttt{GET}            \\ \hline
    \multirow{4}{*}{\begin{tabular}[c]{@{}c@{}}\textbf{Training}\\  \textbf{Accuracy}\\ ($R^2$, \%)\end{tabular}} &
      2 &
      47.26 &
      49.85 &
      33.22 &
      46.68 &
      49.27 &
      39.37 &
      66.47 &
      \textbf{71.80} \\
     & 4  & 60.90 & 62.90   & 54.62    & 62.59  & 55.81     & 53.34    & 79.51          & \textbf{82.40} \\
     & 8  & 81.33 & 81.24   & 78.12    & 82.28  & 66.41     & 64.70    & 90.81          & \textbf{91.02} \\
     & 12 & 93.43 & 92.97   & 93.48    & 94.90  & 72.93     & 66.83    & \textbf{97.50} & 96.09          \\ \hline
    \multirow{4}{*}{\begin{tabular}[c]{@{}c@{}}Testing\\  Accuracy\\ ($R^2$, \%)\end{tabular}} &
      2 &
      46.45 &
      48.15 &
      32.86 &
      46.12 &
      48.59 &
      38.45 &
      64.44 &
      70.24 \\
     & 4  & 58.60 & 59.76   & 53.14    & 61.24  & 55.27     & 51.40    & 74.84          & 77.89          \\
     & 8  & 69.07 & 68.26   & 70.16    & 74.18  & 66.03     & 62.39    & 74.00          & 78.55          \\
     & 12 & 67.21 & 64.27   & 63.80    & 67.63  & 72.32     & 63.67    & 67.14          & 71.97          \\ \hline
    \multirow{4}{*}{\begin{tabular}[c]{@{}c@{}}Training\\  Time\\ ($s$)\end{tabular}} &
      2 &
      0.03 &
      2648.84 &
      0.59 &
      3.47 &
      1054.35 &
      28.74 &
      313.89 &
      796.65 \\
     & 4  & 0.04  & 3439.93 & 1.18     & 5.72   & 1680.10   & 49.24    & 673.39         & 2234.64        \\
     & 8  & 0.06  & 3932.06 & 3.46     & 8.87   & 10415.84  & 277.17   & 7872.43        & 2420.84        \\
     & 12 & 0.08  & 4179.08 & 9.12     & 15.70  & 173544.28 & 9564.43  & 181308.67      & 9394.67        \\ \bottomrule
    \end{tabular}%
    }
    \end{table}

\textbf{The effectiveness of our tree optimization approach:} \cref{tab:TrainforAllMethods} shows that our method \texttt{GET} outperforms \texttt{ORT-LS} by 5.33\%, 2.89\% and 0.21\% in training accuracy for depths of 2, 4 and 8, respectively, while it outperforms \texttt{CART} by 24.55\%, 21.50\% and 9.69\% across various depths. These improvements in training accuracy verify the efficacy of our tree optimization method to achieve training optimality. Additionally, an increase in training accuracy correlates with improved testing accuracy at depths of 2, 4, and 8, suggesting that an optimized tree with higher training accuracy can potentially yield better testing accuracy before encountering serious overfitting issues. Overfitting, particularly at deeper depth like 12, is simply addressed by tuning an optimal tree depth through cross validation, as discussed in previous subsections.    

\textbf{Ablation study on the strategies of our optimization approach:} 
The observed improvements in training accuracy of our method \texttt{GET}, as reported in \cref{tab:TrainforAllMethods}, can be attributed to two key strategies: the iterative scaled sigmoid approximation and the subtree polish strategy. The comparative results, presented in \cref{tab:ablationStrategy}, illustrate the impact of these strategies on \texttt{GET}. For clarity, in the \cref{tab:ablationStrategy}, our method that utilizes the iterative scaled sigmoid approximation, is referred to as \texttt{GET}. In contrast, when this strategy is not employed, the method is denoted by specific values of scale factors, such as $\alpha=1$ and $\alpha=100$. Additionally, the subtree polish strategy is evaluated in the table, with the methods labeled as ``\texttt{GET without subtree polish strategy}'' and ``\texttt{GET with subtree polish strategy}''.

\begin{table}[!htp]
    \centering
    \caption{The effectiveness of strategies used in our tree optimization approach on training accuracy.}
    \label{tab:ablationStrategy}
    \resizebox{\columnwidth}{!}{%
    \begin{tabular}{cc|cccc}
    \toprule
    \multicolumn{2}{c|}{Item}                                & $D=2$ & $D=4$ & $D=8$ & $D=12$ \\ \hline
    \multicolumn{1}{c|}{\multirow{2}{*}{\begin{tabular}[c]{@{}c@{}}without   Iterative Scaled Sigmoid Approximation \\      (A fixed scale factor)\end{tabular}}} &
      \begin{tabular}[c]{@{}c@{}}$\alpha=1$ \\ (Sigmoid Function)\end{tabular} &
      55.23 &
      65.38 &
      81.91 &
      93.42 \\ \cline{2-6} 
    \multicolumn{1}{c|}{} & $\alpha=100$                      & 68.48 & 77.22 & 87.09 & 94.11  \\ \hline
    \multicolumn{1}{c|}{\multirow{2}{*}{\begin{tabular}[c]{@{}c@{}}with   Iterative Scaled Sigmoid Approximation \\      (Our \texttt{GET} method)\end{tabular}}} &
      \texttt{GET} without subtree polish strategy &
      70.86 &
      80.30 &
      89.45 &
      95.16 \\ \cline{2-6} 
    \multicolumn{1}{c|}{} & \texttt{GET} with subtree polish strategy & 71.80 & 82.40 & 91.02 & 96.09  \\ \bottomrule
    \end{tabular}%
    }
    \end{table}

Regarding the iterative scaled sigmoid approximation, we analyze training accuracy with and without this strategy. Without this strategy, we utilized a fixed scale factor for the differentiability approximation, utilizing both the standard sigmoid function with $\alpha=1$ and a larger scale factor with $\alpha=100$. The findings indicate substantial improvements with the iterative approach: training accuracy increased by 15.63\%, 14.92\%, 7.54\%, and 1.74\% at tree depths of 2, 4, 8, and 12, respectively, when compared to the standard sigmoid approximation. Moreover, when compared to the larger scale factor of $\alpha=100$, the iterative approach improved training accuracy by 2.38\%, 3.08\%, 2.36\%, and 1.05\% for these depths, respectively. These results confirm that our iterative scaled sigmoid approximation surpasses both the standard sigmoid function commonly used in soft decision trees and larger scale factors in differentiability approximation.

Additionally, the subtree polish strategy contributes an additional improvement, particularly at shallower tree depths. It boosts training accuracy by 0.94\%, 2.10\%, 1.57\%, and 0.93\% at tree depths of 2, 4, 8, and 12, respectively.

\textbf{Training time analysis:} A detailed training time comparison is given in \cref{tab:TrainforAllMethods}. Our method \texttt{GET} not only significantly improves training accuracy but also shows great scalability over the state-of-the-art \texttt{ORT-LS} by approximately 20 times acceleration at deep depth 12. However, it remains considerably slower, thousands of times, than \texttt{CART}. This trade-off between training optimality and computational cost is deemed acceptable for a single tree model. In comparison with random forests, our \texttt{GET} offers competitive accuracy but incurs substantially longer training times, being thousands of times slower than \texttt{RF}. Despite this inefficiency, our focus mainly lies in making a single tree suitable for scenarios with limited computational resources. Consequently, our primary metrics of interest are testing accuracy, interpretability and testing time, rather than training time. The benefits of its lightweight structure and prediction speed are further discussed in \cref{subsec:predictionTimeCompr}.

\subsection{Analysis for Parameter Number, Prediction Time, and Interpretability}
\label{subsec:predictionTimeCompr}
As discussed in \cref{sec:intro}, random forests usually replace a single decision tree to improve testing accuracy in embedded systems. However, a single tree with comparable accuracy could be preferable due to its lightweight structure and interpretability. In this section, we primarily compare our methods with \texttt{RF} regarding the number of parameters, prediction time, and interpretability.

\textbf{Comparison of parameter number and testing time:} Building on the testing comparison in \cref{tab:TestVSRF}, we then assess total parameter number and prediction time, in \cref{tab:predictiontime}. \texttt{RF} contains 324 times more parameters than \texttt{GET} and 119 times more than \texttt{GET-Linear}. The prediction speed of \texttt{GET}, averaged over 10,000 repetitions, is 30 times faster than \texttt{RF}, and \texttt{GET-Linear} is 24 times faster. This comparison shows that our tree achieves competitive accuracy with significantly fewer parameters and faster prediction than \texttt{RF}, thereby saving memory and computational costs.

\begin{table}[!htp]
    \centering
    \caption{Comparison of parameter number and prediction time for \texttt{GET}, \texttt{GET-Linear}, and \texttt{RF}.}
    \label{tab:predictiontime}
    \resizebox{\columnwidth}{!}{%
    \begin{tabular}{c|cccc|c|c}
    \toprule
    Item &
      \begin{tabular}[c]{@{}c@{}} Number \\     of Branch Nodes\end{tabular} &
      \begin{tabular}[c]{@{}c@{}} Number \\      of Leaf Nodes\end{tabular} &
      \begin{tabular}[c]{@{}c@{}} Parameter Number \\      in Branch Nodes\end{tabular} &
      \begin{tabular}[c]{@{}c@{}} Parameter Number \\      in Leaf Nodes\end{tabular} &
      Total Parameters &
      \begin{tabular}[c]{@{}c@{}}Prediction Time \\      (s)\end{tabular} \\ \hline
    \texttt{RF}         & 819,566.56 & 819,875.94 & 1,639,133.13 & 819,875.94 & 2,459,009.06 & 1.7337 \\
    \texttt{GET}        & 476.50    & 477.50    & 7,101.63    & 477.50    & 7,579.13    & 0.0572 \\
    \texttt{GET-Linear} & 1,084.25   & 1,085.25   & 10,292.13   & 10,304.50  & 20,596.63   & 0.0728 \\ \bottomrule
    \end{tabular}%
    }
\end{table}

\textbf{Interpretability of our oblique tree:} Interpretability can be assessed from two aspects: tree-based prediction logic and the complexity of decision rules. First, \texttt{RF}, in \cref{tab:TestVSRF}, uses an average of 309.38 trees for higher accuracy, almost losing the interpretability for final predictions compared to a single tree. Second, \texttt{RF} often results in an overly-deep tree with an average depth of 19.24, significantly deeper than our \texttt{GET} with depth of 6.56. Moreover, as compared in \cref{tab:testVSDT}, \texttt{GET} not only achieves the highest testing accuracy but also with the lowest tree depth, enhancing interpretability. For instance, a 2-depth tree yields 4 decision rules across 2 layers, whereas 8-depth tree produces 256 rules across 8 layers. Understanding hundreds of nested IF-THEN rules can be challenging. Therefore, our optimized tree with smaller depth offers more interpretability than deeper decision trees.
% Smaller tree depth typically yields higher degree of interpretability, as it generates fewer decision rules associated with final predictions.

\subsection{Limitations Analysis of Our Approach}
\label{subsec:limitations}
Our extensive experiments are conducted on datasets containing less than 20,000 samples, without exploring large-scale datasets. While our approach exhibits great scalability compared to the state-of-the-art heuristic decision tree \texttt{ORT-LS}, it is still thousands of times slower than the heuristic methods \texttt{CART} and \texttt{RF}. This limitation is generally acceptable for a single tree model, particularly because it significantly improves accuracy, aligning with our primary focus on testing accuracy, interpretability, and prediction time. However, the longer training time could potentially limit the application of our method in scenarios involving large-scale datasets. 

Another limitation arises from inadequate regularization in our optimization approach. As shown in \cref{tab:TrainforAllMethods}, while training accuracy for \texttt{GET} improves by 5.07\% from depth 8 to 12, testing accuracy conversely drops by 6.58\%, indicating a serious overfitting issue. This overfitting issue is more pronounced in decision trees with linear predictions, with more detailed analysis and comparisons provided in \cref{appendix:overfitting}. In response, we preliminarily apply $L_1$ regularization to \texttt{GET-Linear} for the experiments reported in \cref{tab:TestVSRF}, leading to a modest improvement in testing accuracy by 0.73\%. However, due to the challenges in identifying the optimal regularization strength and potential increases in computational costs, we limited our tuning to only between 0 and a small value of $1e-5$, without extensive tuning. Despite these efforts, further enhancements in testing accuracy could be achieved through more dedicated regularization strategies.

\section{Conclusion}
\label{sec:conclusion}
In conclusion, our development of the gradient-based entire tree optimization method, is not necessarily to bring the best regression tree, but rather to explore the potential of a single decision tree in achieving comparable testing accuracy to the classic random forest. This makes a single tree more preferable for scenarios where a lightweight structure and interpretability are valued alongside predictive performance. Our approach reformulates decision tree training as a differentiable unconstrained optimization task, incorporating an iterative scaled sigmoid approximation. The tree optimization capability is further enhanced by a subtree polish strategy. Extensive experiments show that our optimized tree not only achieves but also statistically confirms its testing accuracy comparable to classic random forest, challenging the mindset that a single decision tree typically underperforms random forests in testing performance.

Our study does not imply that base tree models are superior to ensemble models. Rather, our optimized tree provides a promising alternative for scenarios where a single decision tree is a perfect target instead of the classic random forest. Furthermore, our proposed approaches can also be extended to forest models using ensemble learning, which we intend to explore in future work.

\clearpage
\bibliography{iclr2025_conference}
\bibliographystyle{iclr2025_conference}
\nocite{*}

\clearpage 
\appendix

\input{appendix.tex}

\end{document}

%% file: math_commands.tex
\usepackage{amsmath,amsfonts,bm}

\def\eqref#1{equation~\ref{#1}}
\def\1{\bm{1}}

\def\va{{\bm{a}}}
\def\vb{{\bm{b}}}

\def\vh{{\bm{h}}}

\def\vk{{\bm{k}}}

\def\vx{{\bm{x}}}

\def\mA{{\bm{A}}}

\def\mK{{\bm{K}}}

\DeclareMathAlphabet{\mathsfit}{\encodingdefault}{\sfdefault}{m}{sl}
\SetMathAlphabet{\mathsfit}{bold}{\encodingdefault}{\sfdefault}{bx}{n}

\def\sA{{\mathbb{A}}}

\def\sR{{\mathbb{R}}}

\def\sT{{\mathbb{T}}}

%% file: appendix.tex
\section{Certain Scenarios Requiring Decision Trees with Hard-splits}
\label{appendix:scenarios}
Hard-split decision trees and soft decision trees represent fundamentally different models, each suited to different application scenarios. Further, there do exist scenarios where the application of hard-split decision trees is not only more appropriate but also imperative. Below are illustrative examples from several
industrial projects that underscore this preference:

Scenario 1: Threshold-based well control optimization in underground hydrogen storage system.

In the context of large-scale underground hydrogen storage, the operation of periodic hydrogen injection and production across numerous wells necessitates the design of optimal control strategies. Such decision-making problems can be modeled by decision trees, particularly when decisions center on exceeding specific
thresholds. For instance, the reservoir pressure or the production rate exceeding a certain threshold can be regarded as the hard decisions at branch nodes. For these scenarios, hard-split decision trees are more suitable for making clear decisions, which can be easily understood and implemented by field engineers or operators. 

Scenario 2: True-false decision-making in law enforcement.

In law enforcement, decision-making often involves binary choices, such as whether to prosecute a suspect or not. In this case, hard-split decision trees can be used to make clear decisions based on the evidence and the law, which can be easily interpreted by legal professionals.

Scenario 3: Piece-wise affine control law in explicit model predictive control.

In explicit model predictive control, the control law is often represented by piece-wise affine functions, which can be effectively approximated by a hard-split decision tree. The hard decisions at branch nodes can be used to determine the control action based on the state of the system, which can be easily implemented
in real-time control systems.

Beyond the utility of making clear hard decisions, the optimal decision tree model that offers superior predictive performance is also crucial for these scenarios. From the perspective of application scenarios, the motivation and necessity for hard-split optimal decision trees are both intuitive and compelling.

\section{Empirical Analysis of The Impact of Scale Factors on Scaled Sigmoid Approximation}
\label{appendix:alpha}
The scale factor $\alpha$ significantly influence the approximation degree to indicator function and the behavior of its gradient in optimization. A larger $\alpha$ leads to a better approximation than sigmoid function, achieving closer proximity to an indicator function as $\alpha$ approaches infinity. Nonetheless, a larger $\alpha$ also results in a more unstable gradient, which may adversely affect the optimization process. Specifically, a larger $\alpha$ may cause the gradient to be too small or even zero, leading to the stagnation of gradient descent. 

The gradient of scaled sigmoid function, denoted as $S(x)=(1+e^{-\alpha x})^{-1}$ is given by $\frac{\partial{S}}{\partial{x}} = \alpha S(x)(1-S(x))$. When $\alpha$
is larger, either $S(x)$ or $1-S(x)$ tend towards zero for value of $x$ away from origin, potentially causing the gradient to approach zero.
Consequently, the $\alpha$ represents a critical balance between achieving high approximation degree and maintaining stability in optimization processes.

To show how the scale factor $\alpha$ impacts the training optimality of the unconstrained optimization problem (our method without using the strategy of iterative scaled sigmoid approximation), we conduct the comparison experiments across different $\alpha$ values at different depths. The findings, summarized in \cref{tab:alpha}, reveal the relationship between $\alpha$ and training performance, as measured by the average training accuracy $R^2$. Notably, we observe that the training accuracy at the extremes of $\alpha=1$ (sigmoid function) and $\alpha=1000$ are inferior compared to intermediate $\alpha$ values. This observation underscores two critical insights: firstly, relying solely on the sigmoid function ($\alpha=1$) yields suboptimal optimization results; secondly, the high $\alpha$ value may not necessarily lead to better optimality.

\begin{table}[!htp]
  \tabcolsep=22pt
  \centering
  \caption{The impact of $\alpha$ on training accuracy across different depths.}
  \label{tab:alpha}
  \resizebox{\columnwidth}{!}{%
  \begin{tabular}{c|cccc}
  \toprule
  \multirow{2}{*}{Various $\alpha$ Value} & \multicolumn{4}{c}{Training Accuracy ($R^2$, \%)} \\ \cline{2-5} 
                                          & $D=2$    & $D=4$    & $D=8$    & $D=12$    \\ \hline
  $\alpha=1$  (Standard Sigmoid Function) & 55.23    & 65.38    & 81.91    & 93.42     \\
  $\alpha=50$                             & 67.97    & 78.72    & 87.51    & 93.99     \\
  $\alpha=100$                            & 68.48    & 77.22    & 87.09    & 94.11     \\
  $\alpha=1000$                           & 65.37    & 75.64    & 83.26    & 93.42     \\ \bottomrule
  \end{tabular}%
  }
  \end{table}

However, it remains a challenge to identify the optimal scale factor $\alpha$ that balances the trade-off between approximation degree and differentiability. To mitigate this issue, we propose an iterative scaled sigmoid approximation strategy, detailed in the following \cref{subsec:iterative}, to narrow the gap between the original indicator function and its differentiable approximation.

\section{Deterministic Calculations for Leaf Prediction Parameters}
\label{appendix:deterministicCal}
As outlined in \cref{alg:GET}, our method deterministically calculates the leaf values (i.e., the values of $\mK$ and $\vh$), rather than directly using trained values for $\mK$ and $\vh$. Given a tree with tree split parameters $\mA$ and $\vb$, the deterministic tree path for each sample can be calculated by \cref{eq:pt}, which allows for determining the total number of samples assigned to a specific leaf node $t\in\sT_L$. 

For decision trees with constant predictions, the value of $\mK$ remains zero. The value of $\vh$ at a leaf node $t$ is an average of true output values ($y_i$) of the samples assigned to that leaf node $t$.

For decision trees with linear predictions, the prediction at a leaf node is a linear combination of input features by fitting a linear correlation between all samples assigned to that leaf node. The leaf values at a leaf node $t$, $\vk_t$ and $h_t$, are the linear coefficients determined by linear regression.

\section{Hyperparameters Analysis of Our Gradient-based Entire Tree Optimization Approach}
\label{appendix:hyperparameters}
Despite the introduction of additional hyperparameters in gradient-based optimization, tuning them is not typically necessary because their effects are straightforward. To be more specific, the hyperparameters in our gradient-based entire tree optimization approach are as follows:

(1) The multi-start number $N_{start}$

% \begin{figure}[!htbp]
\begin{figure}[!htb]
    % \vskip 0.2in
    \begin{center}
    \centerline{\includegraphics[width=\columnwidth]{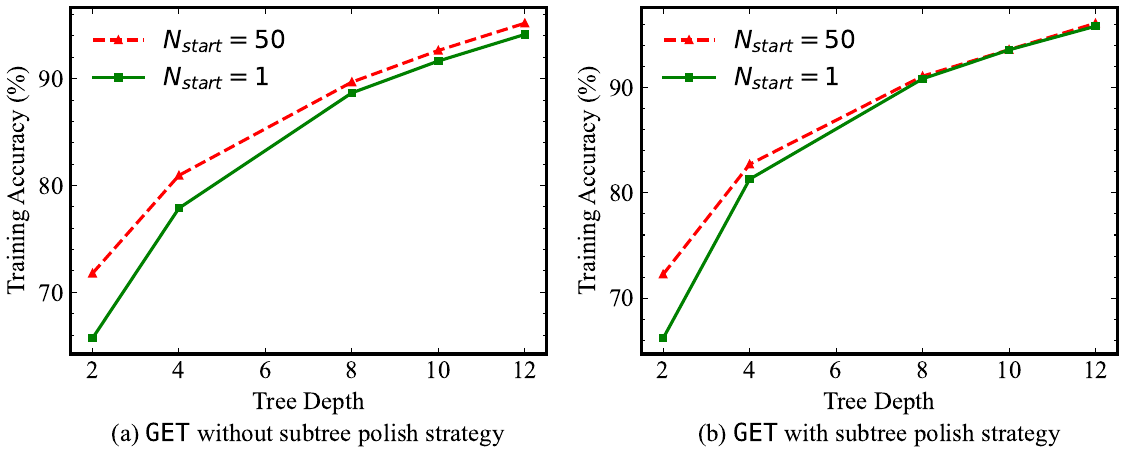}}
    % \vskip -0.1in
    \caption{The trend of training optimality under different depth setting with different $N_{start}$.}
    \label{fig:nstart}
    \end{center}
    % \vskip -0.5in
 \end{figure}
The multi-start number $N_{start}$ directly influences training optimality by increasing the chance of finding the optimal solution, albeit at a higher computational cost. In practice, $N_{start}$ is set to balance acceptable computational expenses with desired training accuracy. 

To explore the correlation between $N_{start}$ and the training optimality, our Gradient-based Entire Tree Optimization for decision trees with constant predictions (\texttt{GET}) is performed under different $N_{start}$ values as shown in \cref{fig:nstart}. It indicates that increasing  $N_{start}$ generally improves training optimality for all various tree depths, especially at lower tree depths.

(2) The epoch number $N_{epoch}$

The epoch number $N_{epoch}$ is another hyperparameter that directly affects training optimality. A higher $N_{epoch}$ value increases training accuracy, but it also increases computational costs. In practice, $N_{epoch}$ is also set to balance acceptable computational expenses with desired training accuracy.

\begin{figure}[!htb]
    % \vskip 0.2in
    \begin{center}
    \centerline{\includegraphics[width=0.7\columnwidth]{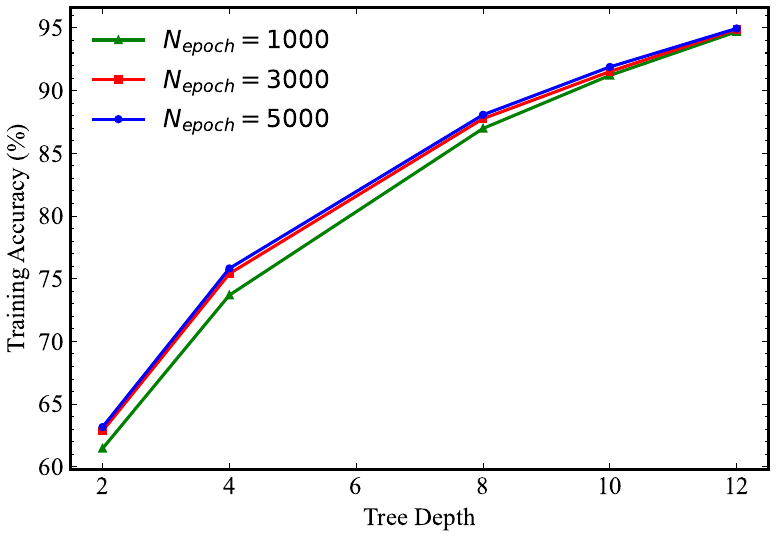}}
    % \vskip -0.1in
    \caption{The trend of training optimality under different depth setting with various $N_{epoch}$.}
    \label{fig:nepoch}
    \end{center}
    % \vskip -0.5in
 \end{figure}

Our experiment with different $N_{epoch} = \{100, 3000, 5000\}$ in \cref{fig:nepoch}, shows that increasing $N_{epoch}$ generally improves training optimality for all various tree depths. The improvements are more pronounced for lower $N_{epoch}$, and become less significant as $N_{epoch}$ increases. 

(3) The range of sampled scale factors $[\alpha_{min}, \alpha_{max}]$

\begin{figure}[!htbp]
    % \vskip 0.2in
    \begin{center}
    \centerline{\includegraphics[width=0.6\columnwidth]{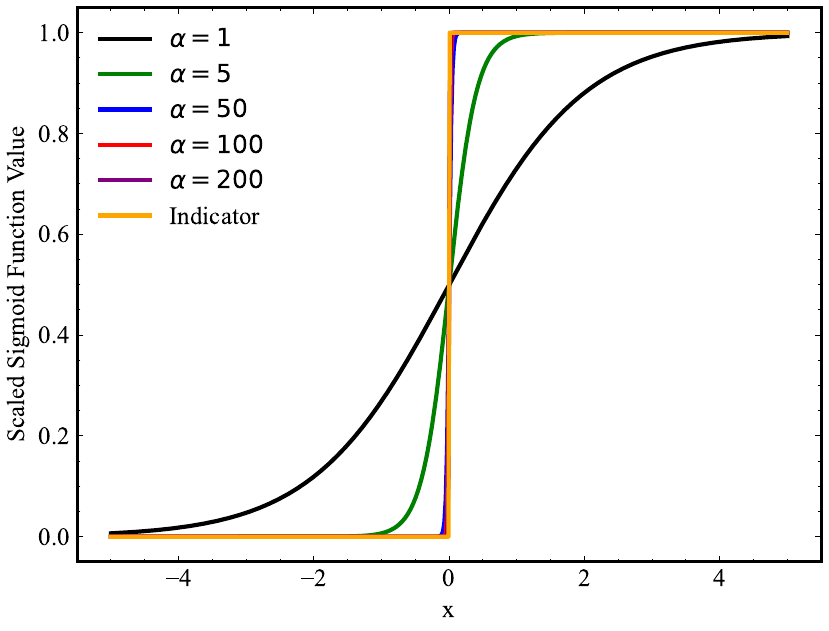}}
    % \vskip -0.1in
    \caption{Disparity comparison of scaled sigmoid function over indicator function under varying $\alpha$.}
    \label{fig:alpha}
    \end{center}
    % \vskip -0.5in
 \end{figure}

This predetermined range is used to sample a set of scaled factors $\alpha$ for the strategy of iterative scaled sigmoid approximation. The principal aim is to explore a broader range of scale factors, ranging from smaller to larger values. As observed in \cref{fig:alpha}, the gap between scaled sigmoid approximation and the original indicator function narrows as $\alpha$ increases. Noticeably, the standard sigmoid function with $\alpha=1$ exhibits a significant deviation from the indicator function as depicted in black line. In the implementation of our experiments, we simply set the range $[\alpha_{min}, \alpha_{max}] = [5,150]$ meet our requirements. This range ensures that smaller values maintain a smooth gradient and exhibit less disparity than the standard sigmoid function, while larger values closely approximate the indicator function.

(4) The number of sampled scale factors 

Within a predetermined range, a set of scaled factors, denoted as $\{\alpha_1, \cdots, \alpha_n\}$, is sampled and sorted in ascending order for subsequent use in iterative scaled sigmoid approximation, as detailed in \cref{subsec:iterative}. Larger scale factors reduce the approximation disparity, whereas smaller ones yield a smoother and more stable gradient. Including a greater number of scale factors in the set facilitates a more stable approximation process, enhancing the approximation degree of the indicator function while minimizing the loss of differentiability typically associated with larger scale factors. Intuitively, including more scale factors in the set enhances training optimality. However, this leads to increased iterations in the approximation strategy, thereby raising computational costs. Practically, the number of scale factors is often determined by balancing training accuracy against computational demands. In our experiments, to avoid excessive computational costs, we primarily sample two scale factors: a smaller $\alpha$ within the range of $[5, 25]$ and larger $\alpha$ within the range of $[50, 150]$.

(5) The learning rate $\eta$ 

The learning rate is a common parameter in gradient-based optimization, and has garnered significant attentions in the literature. To simplify its usage, we adopt the well-established learning rate scheduler, referred to as \texttt{CosineAnnealingWarmRestarts} in PyTorch, which decreases the learning rate from an initial value of 0.01, thus minimizing the need for additional tuning.

\section{Subtree Polish Strategy}
\label{appendix:subtreepolish}
\begin{algorithm}[!htp]
  \small
  \caption{The Subtree Polish Strategy}
  \label{alg:subtreepolish}
  \begin{algorithmic}[1]
     \STATE {\bfseries Input:} Dataset % $(\mathbf{X}$, $\mathbf{y})$,
     $\{\vx_i, y_i\}_{i=1}^n$, tree depth $D$, and other parameters in \cref{alg:GET}.
     \STATE {\bfseries Output:} Optimal trainable variables $\mA_{best}$, $\vb_{best}$, $\mK$ (Zero for the case of \texttt{GET}) and $\vh$.
     \STATE Initially optimize an entire tree as the best tree candidate using \cref{alg:GET}, termed $\mA_{best}$ and $\vb_{best}$. 
     \FOR{$t\in \sT_B$}
        \STATE Induce a subset $\{\mathcal{X}_t, \mathcal{Y}_t \}$ of the dataset $\{\vx_i, y_i\}_{i=1}^n$ for branch node $t$ after fixing the optimized hyperplanes at its parent nodes.
        \IF{$|\mathcal{X}_t|>1$ and $\text{unique}(\mathcal{Y}_t)>1$}
           % \STATE Determine the effective subtree depth $d_{sub}^{ef\!f} = \min\{D-d_t+1, d_{sub}\}$, where $d_t$ is the depth of current node $t$.
           \STATE Retrieve $d_{sub}$-depth subtree results rooted at node $t$ from the tree candidate as the warm start.
           \STATE Polish the subtree, termed as $\mA_{sub}$ and $\vb_{sub}$, using \cref{alg:GET}. \label{line:subtreeopt}
           \STATE Replace the corresponding subtree of the current best tree candidate with $\mA_{sub}$ and $\vb_{sub}$ from \textit{Line~\ref{line:subtreeopt}}. 
           Update the current best tree candidate only if this modification improves training optimality.
        % \ELSE 
        %    \STATE Assign $\mA[t,:] =\mathbf{0}$ and $\vb[t] =\mathbf{0}$. 
        \ENDIF 
     \ENDFOR
     \STATE Deterministically calculate $\mK$ and $\vh$ based on the final best tree candidate $\mA_{best}$ and $\vb_{best}$. 
     % \STATE \textbf{Return} optimal trainable variables $\mA_{best}$, $\vb_{best}$, $\mK$, and $\vh$.
  \end{algorithmic}
\end{algorithm}

\section{The Basic Setting of Numerical Experiments}
\label{appendix:experiment}

\subsection{Dataset Information}
\label{appendix:dataset}
The 16 real-world dataset from UCI repository \citep{Dua_2019_UCI} and OpenML \citep{Vanschoren_2014_OpenML} are used in our numerical experiments. Detailed information about these datasets is summarized in \cref{tab:dataset}. The dataset size $n$ and the number of features $p$ are provided in the table.

\begin{table}[!htp]
    \centering
    \caption{Real-world datasets from UCI and OpenML Repository.}
    \label{tab:dataset}
    \resizebox{\columnwidth}{!}{%
    \begin{tabular}{cccc}
    \toprule
    Dataset   Index & Dataset Name                            & Dataset Size (n) & Feature Number (p) \\ \hline
    1               & airfoil-self-noise                      & 1,503             & 5                  \\
    2               & space-ga                                & 3,107             & 6                  \\
    3               & abalone                                 & 4,177             & 8                  \\
    4               & gas-turbine-co-emission-2015            & 7,384             & 9                  \\
    5               & gas-turbine-nox-emission-2015           & 7,384             & 9                  \\
    6               & puma8NH                                 & 8,192             & 8                  \\
    7               & cpu-act                                 & 8,192             & 21                 \\
    8               & cpu-small                               & 8,192             & 12                 \\
    9               & kin8nm                                  & 8,192             & 8                  \\
    10              & delta-elevators                         & 9,517             & 6                  \\
    11              & combined-cycle-power-plant              & 9,568             & 4                  \\
    12              & electrical-grid-stability               & 10,000            & 12                 \\
    13              & condition-based-maintenance\_compressor & 11,934            & 16                 \\
    14              & condition-based-maintenance\_turbine    & 11,934            & 16                 \\
    15              & ailerons                                & 13,750            & 40                 \\
    16              & elevators                               & 16,599            & 18                 \\    \bottomrule 
    \end{tabular}%
    }
\end{table}

Typically, we allocated 75\% of the samples for training purposes and the remaining 25\% for testing. If an experiment requires cross validation for hyperparameters tuning like tree depth, we then subdivide the training datasets into training and validation subsets in a 2:1 ratio. The dataset setting accordingly changes to 50\% samples as training set, 25\% samples as validation set, and 25\% samples as testing set. After determine the best hyperparameters, we then retrain the model using the combined training and validation set, and use the remaining 25\% as testing set to evaluate the final testing accuracy.

\subsection{The Implementation Settings for Comparison Studies}
\label{appendix:implementation}

To implement our Gradient-based Entire Tree optimization framework, we utilize PyTorch that embeds auto differentiation tools and gradient-based optimizers. Our tree induction method is referred to as \texttt{GET} when applied to trees with constant predictions. In cases of trees with linear predictions, we refer to it as \texttt{GET-Linear}. It should be noted that \texttt{GET} is incorporated our additional subtree polish strategy to mitigate accumulated approximation errors, thereby improving training optimality. In contrast, \texttt{GET-Linear} is not equipped with this strategy, as decision trees with linear predictions generally perform well as observed in our experiments. Moreover, decision trees with linear predictions are prone to overfitting, a risk potentially exacerbated by subtree polish strategy unless carefully regularized. The tendency of these tree to overfit is also evidenced by results from exiting open-source software for model linear trees, discussed in \cref{appendix:overfitting}. Our methods (\texttt{GET} and \texttt{GET-Linear}) are configured with $N_{epoch}=3000$ and $N_{start}=10$, unless otherwise specified.

For benchmarking, the open-source \texttt{Scikit-learn} library in \texttt{Python} is used to implement \texttt{CART} and random forest (\texttt{RF}) methods. The parameter values for these methods are set to default values, unless otherwise specified, such as the specific hyperparameters tuning discussed in \cref{subsec:ComprVSRF}. The implementation of \texttt{HHCART}, \texttt{RandCART} and \texttt{OC1} are adapted from publicly sourced GitHub repository and programmed in \texttt{Python}. We modified their classification-oriented loss functions to adapt for regression tasks. As for the local search method \texttt{ORT-LS}, we reproduce it in \texttt{Julia} due to the absence of open-source code for \texttt{ORT-LS}. The \texttt{GradTree} and \texttt{SoftDT} method are implemented using their respective open-source GitHub repositories, with adjustments made only to the epoch numbers to align with our methods.

Experiments necessitating CPU computation were executed on the high-performance Oracle HPC Cluster, specifically utilizing the ``BM.Standard.E4.128'' configuration. Each compute node within this cluster is equipped with an ``AMD EPYC 7J13 64-Core Processor''. Concurrently, experiments requiring GPU resources were conducted on the 
``Narval'' server, which is equipped with an NVIDIA A100 GPU. Additionally, for the comparative analysis of prediction times as elaborated in \cref{subsec:predictionTimeCompr}, we assessed the prediction speeds of each method on the login node of the ``Cedar'' server.

\subsection{Detailed Test Accuracy Comparison Results on 16 Real-world Datasets}
\label{appendix:testComparison}
% To comprehensively showcase the efficacy of our Gradient-based Entire Tree Optimization approach, we compare our optimized tree \texttt{GET} and \texttt{GET-Linear} with different decision tree methods and random forest \texttt{RF}. 
The testing accuracy ($R^2$) comparison for specific 16 real-world datasets is detailed in \cref{tab:TestEachDataset}. 
\begin{table}[!htp]
    \centering
    \caption{The testing accuracy comparison for each dataset.}
    \label{tab:TestEachDataset}
    \resizebox{\columnwidth}{!}{%
    \begin{tabular}{cccccccc|cc|c}
    \toprule
    Dataset & \texttt{CART}  & \texttt{OC1}   & \texttt{RandCART} & \texttt{HHCART} & \texttt{SoftDT} & \texttt{GradTree} & \texttt{ORT-LS} & \texttt{GET}   & \texttt{GET-Linear} & \texttt{RF}    \\ \hline
    1             & 85.27 & 86.38 & 76.79    & 85.71  & 67.06  & 56.46  & 85.24  & 89.21 & 89.96      & 92.58 \\
    2             & 42.14 & 40.53 & 50.70    & 49.05  & 47.46  & 30.47  & 49.51  & 62.07 & 62.07      & 53.22 \\
    3             & 47.29 & 46.15 & 46.94    & 54.75  & 55.28  & 49.68  & 54.20  & 57.29 & 60.24      & 57.64 \\
    4             & 66.50 & 55.39 & 57.24    & 60.61  & 60.80  & 64.07  & 55.62  & 63.38 & 71.59      & 68.10 \\
    5             & 82.19 & 83.20 & 83.30    & 84.43  & 80.81  & 67.03  & 86.39  & 87.97 & 86.79      & 91.00 \\
    6             & 62.36 & 62.71 & 41.64    & 66.95  & 60.31  & 61.53  & 63.68  & 63.68 & 68.12      & 68.38 \\
    7             & 96.98 & 97.23 & 92.06    & 97.15  & 85.36  & 95.30  & 97.59  & 98.01 & 98.23      & 98.27 \\
    8             & 95.89 & 96.25 & 95.78    & 96.25  & 88.66  & 93.69  & 96.65  & 96.93 & 97.06      & 97.63 \\
    9             & 42.56 & 51.53 & 51.83    & 56.32  & 71.60  & 44.22  & 69.57  & 79.82 & 86.67      & 70.48 \\
    10            & 60.19 & 59.03 & 58.42    & 60.91  & 62.76  & 54.76  & 58.52  & 61.66 & 64.26      & 63.16 \\
    11            & 93.33 & 93.15 & 93.11    & 93.55  & 93.16  & 84.66  & 92.84  & 93.86 & 94.04      & 95.92 \\
    12            & 71.17 & 74.72 & 58.08    & 68.80  & 86.35  & 45.01  & 80.66  & 86.76 & 91.19      & 89.59 \\
    13            & 98.58 & 94.37 & 98.45    & 98.63  & 86.83  & 85.17  & 98.93  & 98.98 & 99.98      & 99.50 \\
    14            & 97.34 & 71.14 & 96.11    & 95.23  & 48.37  & 76.81  & 97.78  & 98.11 & 99.97      & 98.74 \\
    15            & 75.96 & 76.55 & 75.48    & 77.92  & 81.61  & 67.49  & 78.44  & 81.18 & 82.53      & 83.24 \\
    16            & 69.12 & 72.30 & 65.03    & 75.66  & 86.57  & 49.73  & 82.61  & 89.41 & 90.88      & 83.59 \\ \bottomrule
    \end{tabular}%
    }
    \end{table}

\subsection{Overfitting Issue and Comparison for Trees with Linear Predictions}
\label{appendix:overfitting}
As discussed in \cref{subsec:limitations}, overfitting issues has been observed with both our method \texttt{GET} and \texttt{GET-Linear}. Upon further comparison of our \texttt{GET-Linear} with the existing open-source library \texttt{linear-tree}, it is evident that the overfitting issues are more pronounced in trees with linear predictions. The open-source software \texttt{linear-tree} exhibits significantly more severe overfitting issues at depths such as 8 and 12, as detailed in \cref{tab:overfittingGetlinear}. 

For our method \texttt{GET-Linear}, although it benefits from the subtree polish strategy, improving training accuracy by 1\% to 2.19\%, it conversely results in a reduction of testing accuracy by up to 3.24\%. This decrease in testing accuracy for \texttt{GET-Linear} begins at relatively small depths, such as 4. Given the susceptibility of trees with linear predictions to overfitting, we opt not to apply the subtree polish strategy to \texttt{GET-Linear} in order to mitigate the risk of exacerbating overfitting issues. 

\begin{table}[!htp]
    \centering
    \caption{Comparison of training and testing accuracy for \texttt{GET-Linear} and \texttt{linear-tree}.}
    \label{tab:overfittingGetlinear}
    \resizebox{\columnwidth}{!}{%
    \begin{tabular}{c|ccc|ccc}
    \toprule
    \multirow{2}{*}{Depth} & \multicolumn{3}{c|}{Training   Accuracy ($R^2$, \%)}    & \multicolumn{3}{c}{Testing Accuracy ($R^2$, \%)} \\ \cline{2-7} 
     &
      \texttt{linear-tree} &
      \begin{tabular}[c]{@{}c@{}}\texttt{GET-Linear} without \\      subtree polish strategy\end{tabular} &
      \multicolumn{1}{c|}{\begin{tabular}[c]{@{}c@{}}\texttt{GET-Linear} with\\       subtree polish strategy\end{tabular}} &
      \texttt{linear-tree} &
      \begin{tabular}[c]{@{}c@{}}\texttt{GET-Linear} without \\      subtree polish strategy\end{tabular} &
      \begin{tabular}[c]{@{}c@{}}\texttt{GET-Linear} with\\       subtree polish strategy\end{tabular} \\ \hline
    2                      & 79.55 & 81.74 & \multicolumn{1}{c|}{82.71} & 79.15           & 79.98    & 80.76   \\
    4                      & 84.90 & 84.78 & \multicolumn{1}{c|}{86.97} & 72.26           & 81.86    & 80.82   \\
    8                      & 93.13 & 88.59 & \multicolumn{1}{c|}{89.88} & -400195.15 (overfitting)      & 81.50    & 78.26   \\
    12                     & 98.28 & 90.88    & \multicolumn{1}{c|}{92.74}      & -11993680.84 (overfitting)    &      64.91    &  49.53       \\ \bottomrule
    \end{tabular}%
    }
    \end{table}

To mitigate the overfitting issues for \texttt{GET-Linear}, we preliminarily attempt to apply $L_1$ regularization for trainable variables $\mA$. This approach involves incorporating a regularization term into the loss function $\mathcal{L}$ below, serving to penalize the complexity of the tree structure. The regularized loss is delineated in \cref{eq:l1reg}, where $\lambda$ denotes the regularization strength and $\|\cdot\|_1$ represents the $L_1$ norm.
% \begin{equation}
%        \mathcal{L}_{reg} = \sum_{i=1}^{n}\sum_{t\in\sT_L} P_{i,t} \left(y_i-(\vk_t^T \vx_i + h_t)\right)^2 + \lambda \| \mA \|_1 \label{eq:l1reg}
% \end{equation}
\begin{equation}
       \mathcal{L}_{reg} = \sum_{i=1}^{n}\sum_{t\in\sT_L} P_{i,t} \left(y_i-(\vk_t^T \vx_i + h_t)\right)^2 + \lambda \sum_{t\in\sT_b} \| \va_t \|_1 \label{eq:l1reg}
\end{equation}

\begin{table}[!htp]
    \centering
    \caption{The comparison for \texttt{GET-Linear} with and without regularization across 16 datasets.}
    \label{tab:regGetlinear}
    \resizebox{\columnwidth}{!}{%
    \begin{tabular}{c|cc}
    \toprule
    Item                  & \texttt{GET-Linear} without regularization & \texttt{GET-Linear} with regularization \\ \hline
    Testing Accuracy (\%) & 83.24                             & 83.97                             \\ \bottomrule
    \end{tabular}%
    }
    \end{table}

However, identifying the appropriate regularization strength $\lambda$ proves challenging during our experiments, necessitating extensive hyperparameter tuning. This tuning significantly increase the computational cost and the implementation complexity of our method. Consequently, in our experiment reported in \cref{tab:TestVSRF}, we did not extensive tune this parameter. We only adjust it between 0 and a very small value $1e-5$ to implement a minimal regularization, aiming to enhance testing accuracy without greatly compromising optimization capabilities. With this slight regularization, the testing accuracy of \texttt{GET-Linear} reported in \cref{tab:TestVSRF} is improved by 0.73\% compared to the results without regularization, as shown in \cref{tab:regGetlinear}. Despite the slight improvement, the overfitting issues for \texttt{GET-Linear} still exist, and our preliminary regularization is not sufficient to address this issue. Significant improvements in testing accuracy are achievable through appropriate regularization strategies; however, this requires further exploration and is limited in this paper.